# Diffusion Model Regularized Implicit Neural Representation for CT Metal Artifact Reduction

Jie Wen[1], Chenhe Du[1], Xiao Wang[1], Yuyao Zhang[1,*]

School of Information, ShanghaiTech University, Shanghai 201210, China

[*]Correspondence: zhangyy8@shanghaitech.edu.cn

**Abstract (<250 words)**

Computed tomography (CT) images are often severely corrupted by artifacts in the presence of metals. Existing supervised metal artifact reduction (MAR) approaches suffer from performance instability on known data due to their reliance on limited paired metal-clean data, which limits their clinical applicability. Moreover, existing unsupervised methods face two main challenges: 1) the CT physical geometry is not effectively incorporated into the MAR process to ensure data fidelity; 2) traditional heuristics regularization terms cannot fully capture the abundant prior knowledge available. To overcome these shortcomings, we propose diffusion model regularized implicit neural representation framework for MAR. The implicit neural representation integrates physical constraints and imposes data fidelity, while the pre-trained diffusion model provides prior knowledge to regularize the solution. Experimental results on both simulated and clinical data demonstrate the effectiveness and generalization ability of our method, highlighting its potential to be applied to clinical settings.

**Author summary:** We present INR-DR, an unsupervised framework that combines implicit neural representations and diffusion priors to remove metal artifacts from CT images without retraining. Tested on dental and abdominal scans, it outperforms prior methods and generalizes across unseen metals and geometries.

## 1 INTRODUCTION

Computed tomography (CT) based on X-ray projections has been extensively applied for non-invasive imaging in medical diagnosis and therapy guidance. Reconstructing CT images from X-ray projection data is a typical inverse problem. Regrettably, when metallic implants are present within the body, e.g., dental fillings and vascular stents, X-ray beams that directly pass through them are significantly absorbed. This results in corresponding localized regions of the projections (sinogram domain) becoming unreliable due to the beam hardening effect [1] and often being treated as missing data, making metal artifact reduction (MAR) an ill-posed inverse problem in nature. In this case, directly applying Filtered Back-Projection (FBP) reconstruction can lead to non-local star-shape or streak artifacts. Therefore, the development of MAR methods remains a necessity for CT reconstruction.

Existing methods have demonstrated the effectiveness of replacing missing data with estimated values in the sinogram domain, such as by linear interpolation [2]. To enhance the quality of sinogram interpolation, subsequent approaches incorporate the forward projection of an additional prior image to fill in the missing regions [3]. With the advancement of data-driven deep learning in solving ill-posed inverse problems, recent methods utilize supervised neural networks to further enhance the quality of prior images [4, 5]. However, these sinogram completion methods usually introduce secondary artifacts in reconstructed CT images due to the discontinuity between the estimated filling data and the known sinogram. On the other hand, image-domain networks concentrate on directly learning the transformation from metal-corrupted CT images to their corresponding clean images [6, 7]. These methods usually lack data consistency in handling global artifacts [8].

Recently, several dual-domain networks have been developed to better leverage information from both the sinogram and image domains, achieving better performance compared with single-domain networks [9, 10, 11, 12, 13]. However, most of them perform restoration in both domains separately through two enhancement networks while the physical imaging constraints are not effectively incorporated. Moreover, supervised approaches rely on paired metal-clean data for training, which are usually not available in practice. This also leads to a significant degradation in performance when the metal shapes or the CT scanning protocols are different from those in training data due to the out-of-distribution (OOD) effect.

Implicit neural representation (INR) is an innovative unsupervised deep learning framework for solving ill-posed inverse problems in image reconstruction. It trains a multilayer perceptron (MLP) to represent a single image as a continuous function by mapping coordinates to corresponding pixel values. Hash encoding is also widely incorporated into designed INR networks due to its ability to significantly reduce training complexity [14]. The inherent tendency of MLPs to favor low-frequency signal components serves as an implicit regularization in the image domain. The physical forward model can then be applied to the continuous represented image to produce a corresponding sinogram, which makes INR a promising dual-domain information fusion operator with physical imaging constraints. Therefore, it has achieved significant progress in various CT reconstruction tasks, such as sparse-view CT [15] and limited-angle CT [16]. However, for the MAR task, as data missing in the sinogram domain is more concentrated in certain regions, some localized structural information can be entirely disrupted in CT images, in which case the continuous prior of INR might hold less significance [17, 18]. Therefore, the reconstructed result of INR often loses significant structures like bones, especially in areas around the metal.

With the rise of diffusion models (DMs) [19, 20], it shows potential to introduce generative prior knowledge into ill-posed inverse problems to overcome this shortcoming [21, 22]. These models typically operate by progressively denoising Gaussian noise to produce vivid samples that resemble the target data distribution in a sequence of iterative refinement steps. Given incomplete projection data, a pre-trained DM can sample a prior from estimated posterior probabilities and then help constrain the solution space with different structural constraints over the image at different timesteps. Several approaches have demonstrated the effectiveness of DMs on the MAR task [23, 24, 25, 26], and a common focus is how to effectively embed the measurements to control the generation process towards a data-consistent sample while preserving critical prior knowledge.

Specifically, our contributions can be mainly summarized as:

- We propose Diffusion Model Regularized Implicit Neural Representation (INR-DR), an effective unsupervised framework for the MAR task. Our framework is highly flexible, independent of the type and shape of metals, and compatible with any diffusion model pre-trained on CT images.
- By taking advantage of the unique capability of INR to continuously represent the target image, we can effectively apply the CT forward model to impose physical constraints. Besides, we avoid sinogram stitching, which often leads to secondary artifacts in many other approaches.
- We further utilize the powerful prior provided by unconditionally trained DMs to guide the INR solution space towards the underlying CT image distribution.
- Experiments are conducted on both synthetic metal-affected CT images and clinical data. The results show the effectiveness of our method across various sizes of metallic implants. Furthermore, it demonstrates strong generalization capability even under a cross-body-site clinical setting.

## 2 RESULTS

### 2.1 Baseline and Metrics.

We compare the proposed INR-DR with several current state-of-the-art (SOTA) MAR approaches from various categories, including 1) traditional methods (LI [2] and NMAR [3]); 2) supervised methods (CNN-MAR [5] and InDuDoNet+ [11]); 3) unsupervised diffusion-based methods (DuDoDp [25]).

To ensure fairness, we evaluate the supervised methods using the checkpoints provided by the authors and the diffusion model in the proposed INR-DR is the same as that in DuDoDp. To quantitatively evaluate the MAR performance, we apply commonly-used peak signal-to-noise ratio (PSNR) and structural similarity index measure (SSIM).

Table 1    PSNR/SSIM of different methods on synthesized DeepLesion.

| Method | Large Metal | | | | → Small Metal |
|---|---|---|---|---|---|
| Input | 24.03/0.6436 | 26.78/0.7440 | 30.37/0.7759 | 32.19/0.8136 | 32.92/0.8331 |
| LI [2] | 30.12/0.8785 | 30.82/0.8594 | 35.07/0.9114 | 35.76/0.9134 | 36.18/0.9178 |
| NMAR [3] | 31.54/0.8864 | 31.27/0.8630 | 36.10/0.9177 | 36.37/0.9176 | 36.71/0.9210 |
| CNN-MAR [5] | 32.55/0.9076 | 35.60/0.9119 | 37.07/0.9226 | 37.78/0.9265 | 37.99/0.9330 |
| InDuDoNet+ [11] | <u>37.74</u>/<u>0.9759</u> | <u>40.04</u>/**0.9873** | **43.88**/<u>0.9885</u> | **45.20**/**0.9926** | **46.48**/**0.9927** |
| DuDoDp [25] | 35.39/0.9655 | 35.87/0.9705 | 37.94/0.9762 | 38.22/0.9768 | 38.36/0.9779 |
| INR-DR | **38.28**/**0.9789** | **40.42**/<u>0.9832</u> | 42.98/**0.9892** | <u>44.21</u>/<u>0.9898</u> | <u>44.32</u>/<u>0.9899</u> |

Bold and underline indicate the best and the second best results, respectively.

2.2 Dataset and Preprocessing

We conduct simulation experiments on a well-known public dataset, DeepLesion [27] with metallic implants from [5]. A widely-used simulation protocol [5, 9, 10, 11, 13] is applied to generate synthesized metal-affected CT data, considering the effects of polychromatic X-ray, partial volume effect, beam hardening and Possion noise. All the CT images are converted to linear attenuation coefficient (LAC) maps and resized to 416×416. 640 projection views of fan-beam CT are then uniformly sampled in 360 degrees with 641 detectors for each view, resulting in 640×641 sinograms. Before its integration into our pipeline, the DM has been extensively pre-trained on a total of 927,802 artifact-free CT images from the DeepLesion dataset, the majority of which depict the chest and abdominal regions. Since our INR-DR is an unsupervised learning method in nature, the INR network requires no extra CT images or metal masks for training. For the testing datasets of all methods in our experiment, the data partitioning is consistent with that used in [11,25] (including 200 clean CT images excluded from the DM training dataset and 10 metal masks for testing). The sizes of the 10 metallic implants for test data are 2061, 890, 881, 451, 254, 124, 118, 112, 53, 35 in pixels, which are then divided into five groups to evaluate the artifact-reduction performance across varying implant sizes.

A further experiment is conducted on clinical metal-affected dental CT images from [28]. We follow the preprocessing procedures in [10] to segment metal masks with a threshold of 2500 Hounsfield Units (HU). The metal-affected sinograms and metal traces are then generated by applying the CT forward model to original images and the segmented masks respectively. Since there is no ground truth CT images on clinical data, we only demonstrate the visual results.

2.3 Results on Synthesized DeepLesion

Table 1 shows quantitative results of different MAR methods on the synthesized DeepLesion dataset. To enable a more thorough evaluation of the different methods, we classified the metal implants in the test dataset into five size-based categories, ranging from smallest to largest, and computed the performance metrics individually for each group. In summary, all machine learning-based methods, including our proposed INR-DR, significantly outperform traditional MAR approaches like LI [2] and NMAR [3] in terms of both PSNR and SSIM across all test cases. However, as an earlier supervised learning approach, CNN-MAR[5] employs a relatively simple convolutional neural network to perform inpainting solely in the sinogram domain. Although CNNMAR achieves some improvement over traditional methods, its performance falls short compared to current state-of-the-art techniques such as InDuDoNet+ and DuDoDp, both of which are recently proposed dual-domain restoration methods. Compared with these SOTA approaches, our method demonstrates superior performance across metallic implants of various sizes, with the exception of the supervised method InDuDoNet+. In the cases of medium and small metallic implants, our method maintains a relatively small performance gap with InDuDoNet+, and both methods achieve clinically satisfactory reconstruction metrics.

Firstly, in the case of large metal artifacts, INR-DR achieves the highest PSNR and SSIM values of 38.28 dB and 0.9789, respectively, outperforming all other methods, including the well-designed supervised approach InDuDoNet+ and the diffusion-based unsupervised method DuDoDp. The metrics represent a notable improvement of approximately 0.54 dB in PSNR over InDuDoNet+ and about 2.89 dB over DuDoDp. In terms of SSIM, our method also achieves a notable improvement, outperforming DuDoDp by 0.0134. This demonstrates the competitiveness of our method in more challenging scenarios, effectively eliminating more severe metal artifacts.

For medium-sized metal artifacts, INR-DR demonstrates an even more substantial advantage over DuDoDp, achieving a PSNR improvement of 5.04 dB. However, in this case, InDuDoNet+, as a supervised learning method, demonstrates strong performance on in-domain data, with only a slight gap compared to our approach.

Finally in the scenario of small metal artifacts, INR-DR achieves a satisfactory PSNR of 42.98 dB and an SSIM of 0.9892, closely matching InDuDoNet+ with a minor difference of 0.90 dB in PSNR, while substantially outperforming DuDoDp by around 5.04 dB in PSNR.

Figure 1 also shows the visual comparison on large/mediate/small metals respectively. We will provide a more detailed discussion of the visual results in the Discussion section.

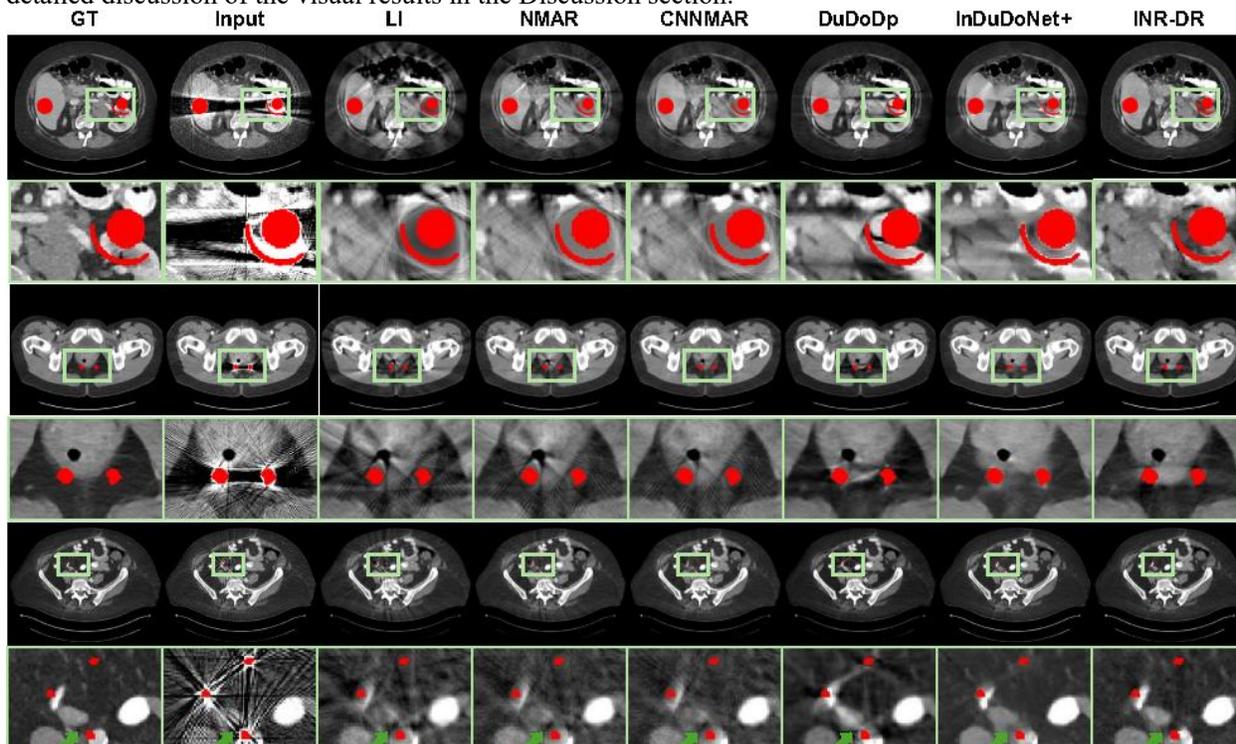

**Figure 1. Comparison on the synthetic DeepLesion dataset.** The display window is [-175, 275] HU and red pixels stand for metallic implants. Row 1,3,5 present an overall visualization of the results for large, medium, and small metallic implants, respectively. Row 2,4,6 provide an enlarged display of the regions with significant differences. In the results for small metallic implants (row 6), green arrows highlight the areas where differences are particularly evident, especially in terms of tissue continuity.

2.4 Results on Clinical Dental CT

To compare generalization ability, we conduct more experiments on clinical dental CT images. Since there is no ground truth for the clinical dataset, we only provide visual comparison instead of metrics. As the metal-included image is from a patient's preoperative diagnosis, we also include postoperative images from the same patient without metal implants as a reference.

As shown in Figure 2, in real-world clinical scenarios, the morphology of metal artifacts often differs significantly from those observed in simulated data. Specifically, the regions affected by beam hardening and photon starvation appear more irregularly distributed, particularly in cases involving multiple metallic implants. This results in the unnatural dark voids observed in the middle of the teeth, as illustrated in the figure. In this case, a significant performance degradation is observed for the result of InDuDoNet+ and DuDoDp, as they both over-suppress non-artifact regions that should clearly retain the original dental structures and introduce more secondary artifacts between the separated metallic implants. INR-DR, however, effectively avoids the generation of secondary artifacts and preserves the structural integrity completely.

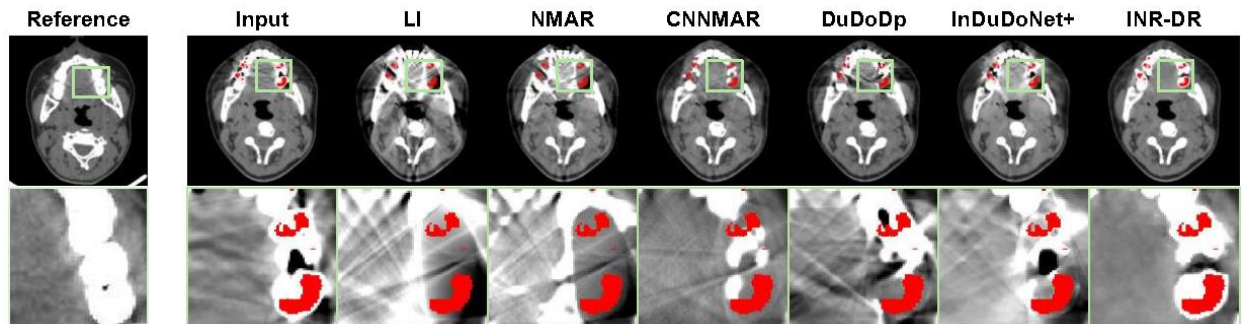

**Figure 2. Comparison on the clinical dental dataset.** The segmentation threshold for metal is 2500HU. The display window is [-175, 275] HU and red pixels stand for the segmented metallic implants. Row 1 presents an overall display of the results. Row 2 provides an enlarged display of the regions with significant differences.

2.5 Reference time and Parameter count

Table 2 reports parameter counts and single-NVIDIA-RTX-4090 runtimes for the evaluated MAR techniques. As end-to-end supervised approaches, both CNNMAR and InDuDoNet+ complete artifact removal in just a few seconds. Although INR-DR is an iterative algorithm and therefore inherently slower than end-to-end CNN-based methods, its inference time remains within three minutes. This clinical-level efficiency is achieved through two critical accelerations: one-step denoising that expedites diffusion inference and hash-based coordinate encoding that markedly accelerates INR convergence, yielding an inference time comparable to that of DuDoDp, thus satisfying the practical constraints of clinical deployment.

In terms of parameter budget, CNN-based approaches exhibit a substantially smaller footprint than the generative DM. Our framework retains the identical diffusion backbone employed by DuDoDp and appends a lightweight INR network equipped with learnable hash encoding, incurring an additional 11.19 million parameters. This parameter count is acceptable for practical use.

**Table 2  Numbers of network parameters and average inference time (seconds) for different MAR methods.**

| Method | CNNMAR [5] | InDuDoNet+ [11] | DuDoDp [25] | INR-DR (Ours) |
|---|---|---|---|---|
| Parameters (M) | 0.03 | 1.75 | 80.11 | 91.30 |
| Inference Time (s) | 0.53 | 1.65 | 42.31 | 97.52 |

# 3 DISCUSSION

3.1 Visual Comparison on Synthesized DeepLesion

In this study, we propose the INR-DR framework for CT metal artifact reduction by combining the unique advantages of INR and diffusion models. Owing to INR's unique advantage of representing an image as a continuous function, we can overcome the resolution limit of the original image and apply the CT forward model more accurately. This approach not only provides a continuity prior for the CT image but also, through precise physical constraints, establishes a strong link between the image and sinogram domains, thereby enhancing the fidelity of the reconstructed results. On the other hand, the pre-trained diffusion model is able to accurately capture the distribution of real CT images. Injecting this distributional prior into the training of the INR network can be regarded as a powerful regularization mechanism, compelling the INR to produce results that are more consistent with the true image distribution.

As an iterative optimization method, INR-DR requires retraining the INR network multiple times during the test phase, resulting in a long runtime. In contrast, end-to-end approaches such as InDuDoNet+ achieve impressively short runtimes. However, a key advantage of our method is that it avoids any sinogram-domain stitching operations, which remain common even in recent state-of-the-art approaches such as InDuDoNet+ [11] and DuDoDp [25]. Most existing approaches in this field treat sinogram regions corrupted by metal artifacts as missing data and subsequently replace them using various interpolation or inpainting techniques. However, such operations often fail to ensure consistency between the filled data and the original uncorrupted measurements, leading to discontinuities in the sinogram domain. These discontinuities can introduce secondary artifacts in the reconstructed images during the back-projection process. Although the dual-domain methods compared in our experiments attempt to mitigate this issue through additional image-domain restoration and achieve excellent results on smaller metals, the problem remains particularly pronounced in cases involving large metal implants, where residual streaking artifacts are still visibly apparent. In contrast, our method employs INR to predict the entire CT image, after which the forward model directly reconstructs a complete and consistent sinogram. Here, data consistency is entirely governed by the INR network, while the prior provided by the diffusion model is injected solely in the form of initial parameters to constrain the INR solution space, thereby preventing the highly uncertain diffusion process from regenerating unreliable artifacts or hallucinations. As a result, even in the presence of large metal implants, our approach achieves superior performance, particularly in eliminating streaking artifacts in the regions connecting metallic objects, as illustrated in Figure 1.

### 3.2 Visual Comparison on clinical dental CT

Moreover, as INR-DR is not a supervised method, it exhibits strong generalization capabilities, independent of the shape, size or type of the testing metals, as well as the geometric parameters and protocols of the CT scan. Although the DM has never been pre-trained on head-specific data and therefore cannot supply accurate structural priors for this out-of-domain scenario, it serves only as a regularizer in our framework. Under such conditions its powerful pre-trained denoising capability can still provide domain-agnostic regularization, implicitly suppressing implausible structures while preserving coherent image features without relying on generative sampling. Moreover, since the INR itself is trained in a fully unsupervised manner without requiring any pre-training data, its representational capacity remains entirely unaffected by the OOD scenario. As a result, the proposed method yields consistent performance on both clinical and synthetic datasets, surpassing purely supervised approaches and iterative schemes that rely more heavily on the DM prior.

For the supervised method InDuDoNet+, however, the training dataset DeepLesion primarily comprises CT images of the chest and abdomen. Besides, the shape and distribution of metallic implants are also more irregular, and the resulting artifacts exhibit more complex and unpredictable patterns in real clinical scenarios. Consequently, the substantial domain gap between the synthetic training data and clinical dental CT images leads to a significant degradation in performance.

### 3.3 Ablation Study

Given that our method proposes the integration of INR and Diffusion models to leverage their respective strengths, we design experiments in this section to selectively ablate the contributions of each component. This approach allows us to elucidate the individual roles of the INR and Diffusion models within our framework and to demonstrate the efficacy of their combination. Furthermore, our ablations confirm that the hash coordinate encoding layer is indispensable for enabling the INR to capture high-frequency image details with markedly fewer parameters.

#### 3.3.1 Effectiveness of Diffusion Prior

To ablate the contribution of the diffusion prior, we train an INR network without diffusion regularization and compare the result with our INR-DR as shown in Figure 3. Here both INR networks have completely the same network settings to ensure fairness. In this testing case, using INR without prior fails to recover critical structures including thoracic vertebrae in these areas while the full-version INR-DR reconstructs these areas properly. In terms of quantitative metrics, the use of diffusion models for regularization results in a PSNR improvement of 2.12 dB and an SSIM increase of 0.0057 compared to the INR-only scenario.

Although INR governs data consistency in the generated results, the prior provided by diffusion is also significant. This is strongly evidenced by our ablation experiments. In the non-convex metal case shown in Figure 3, most of the X-ray beams passing through regions between the two metal parts are strongly absorbed. In our settings, any beam that intersects regions occupied by the metallic implant is classified as corrupted and excluded from the observed sinogram data. As a result, severe data missing is not only limited to the areas surrounding the metallic implants but also includes the enclosed regions, which contain important anatomical structures such as thoracic vertebrae. In this scenario, the information within the affected region is highly underdetermined, resulting in a large solution space for INR that includes anatomically implausible solutions. Generalizing the INR network with the diffusion model effectively excludes these unreasonable solutions, ensuring that the final artifact-reduced result not only maintains data consistency but also preserves anatomically correct structures.

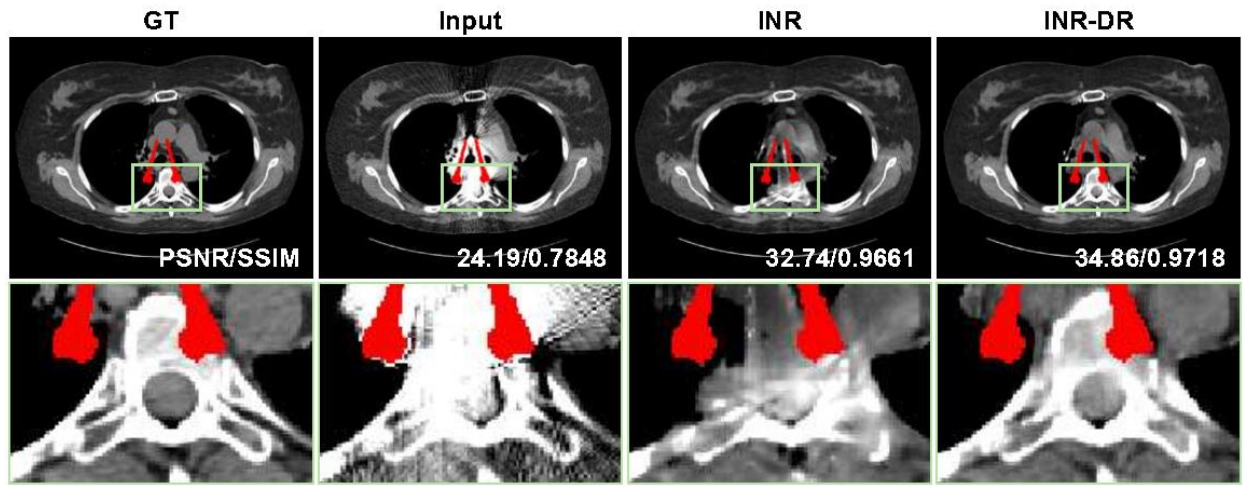

**Figure 3. Comparison of INR without diffusion prior and our full method on synthesized DeepLesion.** The display window is [-175, 275] HU and red pixels stand for the segmented metallic implants. Row 1 presents an overall display of the results. Row 2 provides an enlarged display of the regions with significant differences.

### 3.3.2 Effectiveness of INR

To rigorously examine the indispensability of the INR component in our framework, we conducted a controlled ablation experiment in which the sinogram inpainting was performed solely by the pre-trained DM, completely bypassing the implicit neural representation. The result is compared with the full-version INR-DR as shown in Figure 4. Specifically, we adopted the identical diffusion network that underpins INR-DR, but removed all learnable INR parameters. Starting from a pure noise map, we executed unconditional generation for each diffusion step, produced a synthesized CT image in the image domain, and subsequently applied the discrete CT forward operator to obtain the corresponding synthetic sinogram. The metal trace regions of this synthetic sinogram were then excised and directly pasted into the corrupted regions of the original measured sinogram, following the same blending rule used in INR-DR. After inpainting, the completed sinogram was reconstructed via filtered back-projection to yield a CT slice. This slice was subsequently re-corrupted with the appropriate diffusion schedule noise and fed back into the diffusion network as the initial state for the next denoising step.

Although the DM accurately captures the prior distribution of CT images, its generated samples are not constrained to coincide with the specific observation corresponding to the underlying latent CT slice. Consequently, the inpainted sinogram lacks any mechanism to enforce consistency between the synthesized data within the metal trace and the surrounding measured projections. This discontinuous, "seamed" splice introduces pronounced spectral discrepancies that manifest as severe secondary artifacts after reconstruction, as illustrated in Figure 4 mainly by the dark streaking artifacts along the two bridging metal implants. In addition, the regions adjacent to the metallic insert and those located within the right-sided metallic "gap" suffer from a markedly reduced number of X-ray trajectories that can completely bypass the metals.

Consequently, the corresponding projection data in the sinogram domain are largely discarded as corrupted during standard pre-processing steps. As a result, the attenuation values obtained by direct reconstruction in these regions exhibit pronounced discrepancies relative to their surroundings. When a diffusion model is employed for sinogram inpainting within the iterative framework, the absence of a data-consistency correction permits stochastic deformation of these compromised regions during the sequential re-noising and denoising stages. This leads to the appearance of spurious outliers that propagate into the vicinity of the metallic implant in the final reconstructed image. In contrast, the incorporation of the INR endows the entire CT image with a unified, continuous representation encoded by a single network, thereby obviating any explicit sinogram-level stitching. Consequently, the reconstructed volume remains free of discernible additional artifacts. In terms of quantitative metrics, the integration of INR yields a substantial improvement over DM-only sinogram inpainting, registering a PSNR gain of 6.2 dB and an SSIM increase of 0.0206.

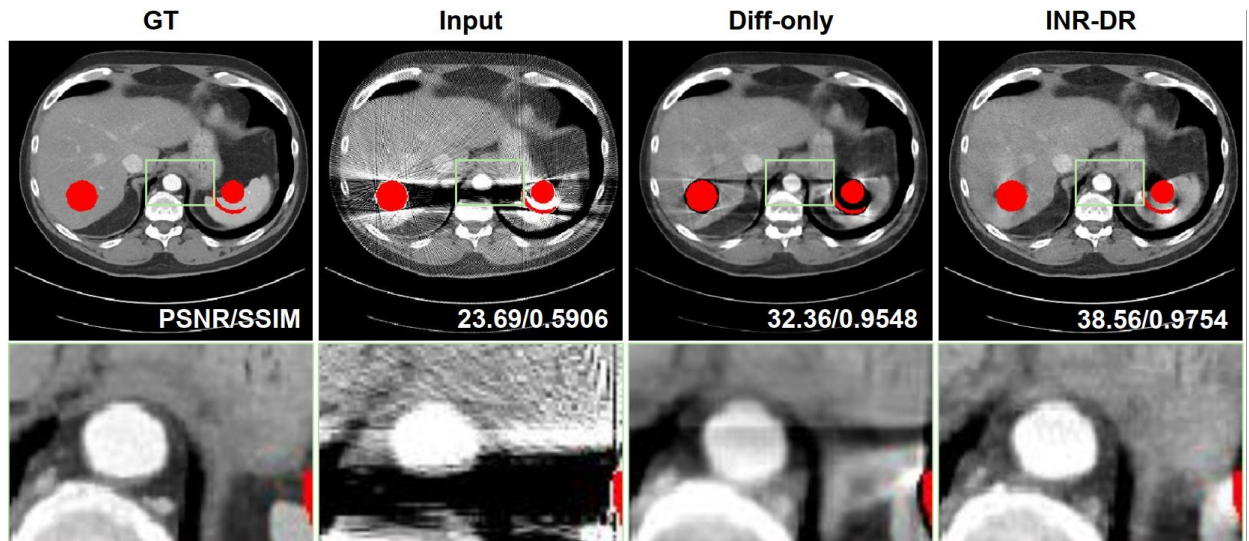

**Figure 4. Comparison of diffusion-inpainting without INR refinement and our full method on synthesized DeepLesion.** The display window is [-175, 275] HU and red pixels stand for the segmented metallic implants. Row 1 presents an overall display of the results. Row 2 provides an enlarged display of the regions with significant differences.

### 3.3.3 Effectiveness of hash coordinate encoding

To validate the necessity of hash-based coordinate encoding, we conducted a controlled ablation in which the INR network was evaluated under three conditions: (i) with no positional encoding at all, (ii) with Fourier (sinusoidal) positional encoding, and (iii) with hash encoding (the full-version INR-DR). The result is demonstrated in Figure 5.

In the absence of any positional encoding, the implicit mapping implemented by a coordinate-based MLP collapses into an inherently low-pass operator. The resulting reconstruction exhibits pronounced spectral roll-off and, perceptually, a conspicuous loss of high-frequency texture that manifests as an overly smooth, blurred image. Fourier (sinusoidal) encoding mitigates this limitation by explicitly projecting the input coordinates into a higher-dimensional feature space spanned by sinusoids of logarithmically increasing frequency, thereby furnishing the network with direct access to high-frequency basis functions. While this strategy successfully restores sharp edges and fine detail, it does so at the cost of uniform spectral allocation: each spatial location is assigned the same, dense set of frequency components regardless of local signal complexity. Therefore, the network expends a substantial fraction of its capacity on smooth regions where high-frequency coefficients are effectively zero, leading to parameter redundancy, slower convergence, and lower data-fidelity in details. Consequently, to merely approach the fidelity depicted in Figure 5 with Fourier encoding, we were compelled to adopt a substantially larger architecture (six hidden layers of 256 neurons each). Unfortunately, the substantial increase in parameter count and the attendant multi-fold slowdown in training speed do not translate into a sufficient gain in fidelity; instead, the reconstructed slice still exhibit excessive blurring of fine anatomical details and a pronounced attenuation of contrast between adjacent tissue types.

In contrast, hash encoding transcends this inefficiency by replacing the fixed sinusoidal basis with a learnable, multi-resolution hash table. The coordinate space is recursively subdivided into voxel grids of exponentially decreasing granularity; a sparse set of latent codes is then retrieved via a deterministic hash function and subsequently decoded by a shallow MLP. This design enables the representation to allocate high-dimensional latent codes only where the local image content exhibits genuine high-frequency energy, thereby achieving a data-adaptive spectral allocation. Consequently, preprocessing the input coordinates with hash encoding enables us to employ an extremely compact network (only two hidden layers of 64 neurons each), yet which reproduces high-frequency image content with superior reconstruction fidelity, as evidenced by the magnified insets in Figure 5. Employing identical network hyper-parameters, Fourier coordinate encoding yields markedly degraded reconstructions in which only gross anatomical contours are retained; the obliteration of fine structural details renders the images unsuitable for reliable clinical diagnosis.

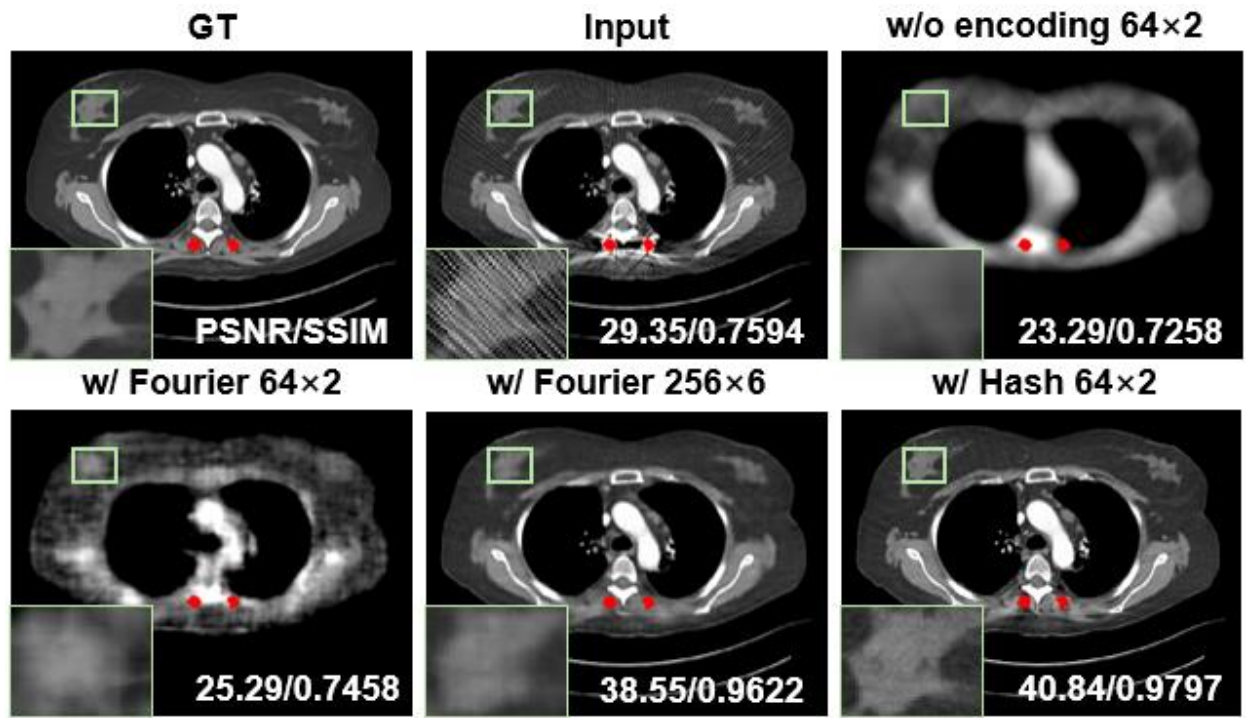

**Figure 5. Comparison of INR-DR with different INR backbone settings on synthesized DeepLesion.** The display window is [-175, 275] HU and red pixels stand for the segmented metallic implants. An enlarged view of the region marked by the green box is shown in the lower-left corner of each image. Notation: "64×2" indicates the network has two hidden layers of 64 neurons each; "256×6" indicates the network has six hidden layers of 256 neurons each; "Fourier" and "Hash" indicate the corresponding coordinate encoding schemes.

### 3.4 Influence of regularization timestep

To effectively inject the distribution prior captured by the diffusion model into the INR, balance the trade-off between data consistency and regularization, and keep the overall training time within a practical limit, the choice of the DM-based regularization timestep $t$ in the one-step denoising process (Eq. 6) is critical. Consequently, we conducted systematic experiments to compare quantitative de-artifacting metrics and total computational cost when $t$ starts from 1000 and is sampled at varying intervals (every 10, 25, 50, 100, 200) across the diffusion trajectory. The results are reported in Figure 6.

Overall, the performance of the proposed framework exhibits a monotonic ascent as the sampling interval decreases, as the extra INR training iterations strengthen data fidelity. This improvement, however, is offset by a corresponding increase in both INR updates and one-step denoising operations, which significantly prolongs total runtime. Empirically, reducing the interval to 50 yields a substantial PSNR improvement and a clear reduction in visual artifacts; beyond this point, further refinement produces only marginal quantitative benefits while runtime grows exponentially, severely undermining clinical feasibility. Therefore, in this paper we set the one-step denoising timestep t 50 steps lower than its previous value each time the INR is regularized by the DM.

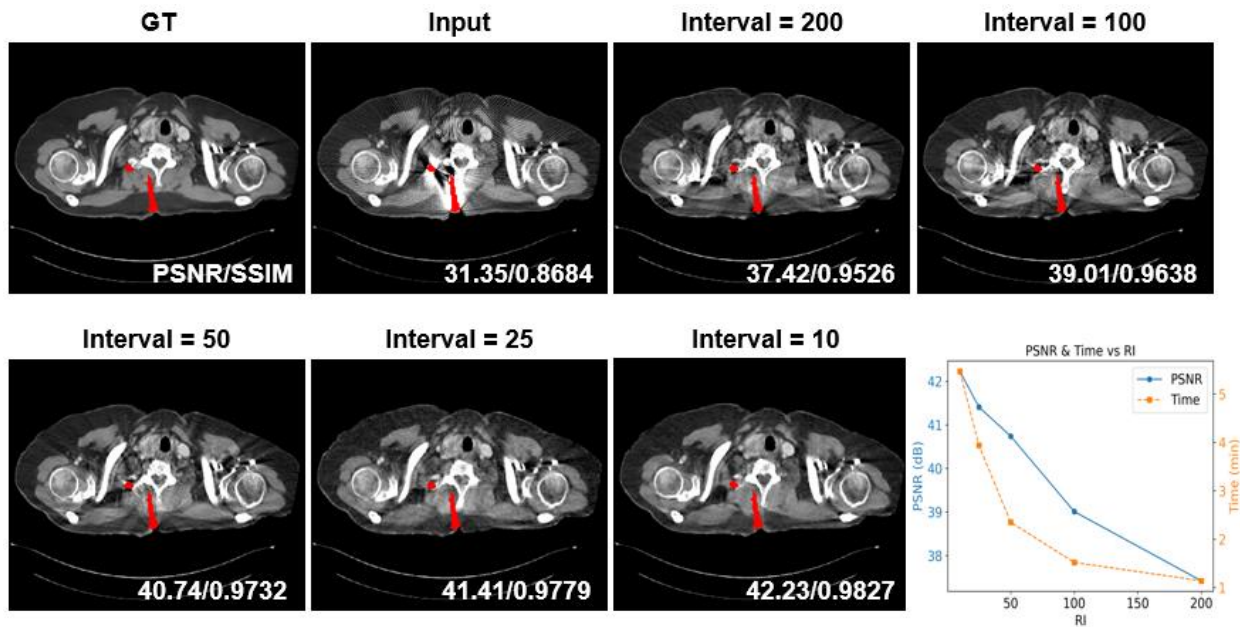

**Figure 6. Qualitative and quantitative comparison of INR-DR with different regularization timestep sampling interval on synthesized DeepLesion.** For visual results, the display window is [-175, 275] HU and red pixels stand for the segmented metallic implants. A dual-axis curve plotting PSNR and total runtime versus the sampling interval is also included to elucidate the performance–efficiency trade-off.

# 4 CONCLUSIONS

This paper presents INR-DR, a novel unsupervised approach for CT metal artifact reduction. Our framework utilizes the implicit neural representation to model the target CT image as a continuous function and then effectively applies the CT forward model to enforce data fidelity with the metal-excluded measurement data. The pre-trained unconditional diffusion model provides prior knowledge to regularize the solution. Experiments on the synthetic DeepLesion dataset and clinical metal-involved dental CT images validate the superior performance and robust generalization ability of our INR-DR, indicating the effectiveness of this unsupervised deep learning approach in the field of MAR.

# 5 MATERIALS AND METHODS

In this section, we present the mathematical modeling of the MAR problem and detail our approach to leverage the strengths of INR and DMs. Specifically, the INR module is dedicated to ensuring data fidelity, and we embed diffusion prior into the training of INR to constrain the solution space. Figure 7 shows a general overview of our approach. Note that the DM used in this paper takes the form of 'Denoising Diffusion Probabilistic Models' (DDPM) [30] and is pre-trained unconditionally on only clean CT images, and we only train the INR network in the proposed pipeline. We always use $\hat{}$ to denote the updated version of the same variable for clarity.

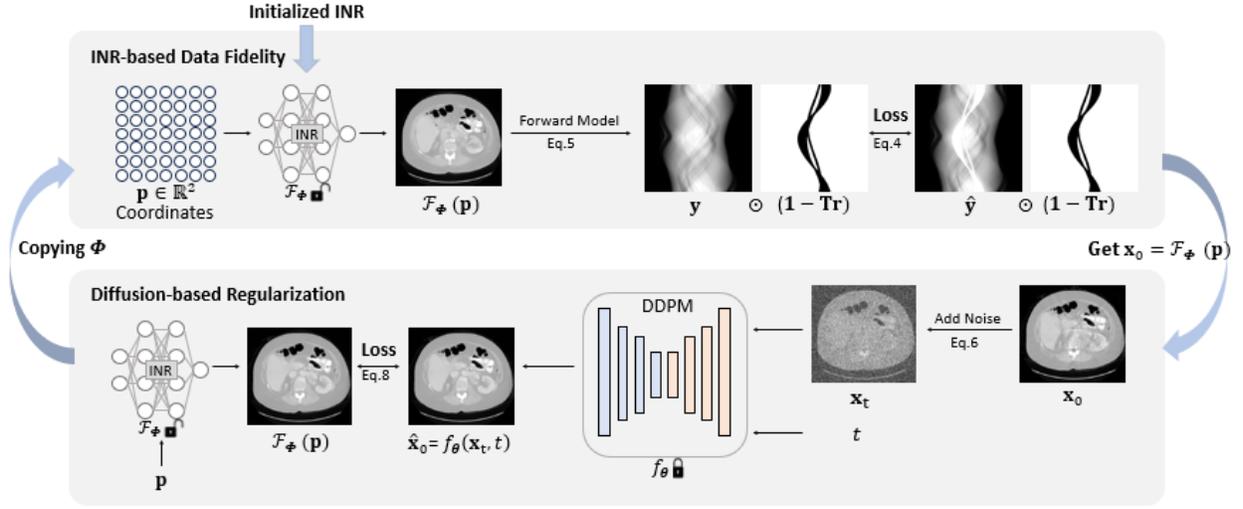

**Figure 7. Overview of the proposed INR-DR.** During one iteration, the INR network undergoes two phases of optimization: (1) Apply the CT forward model to the INR-represented image and enforce data fidelity in metal-affected regions in the sinogram domain; (2) Diffuse and denoise the INR-represented image by the DDPM to obtain a prior image and inject prior knowledge into the INR.

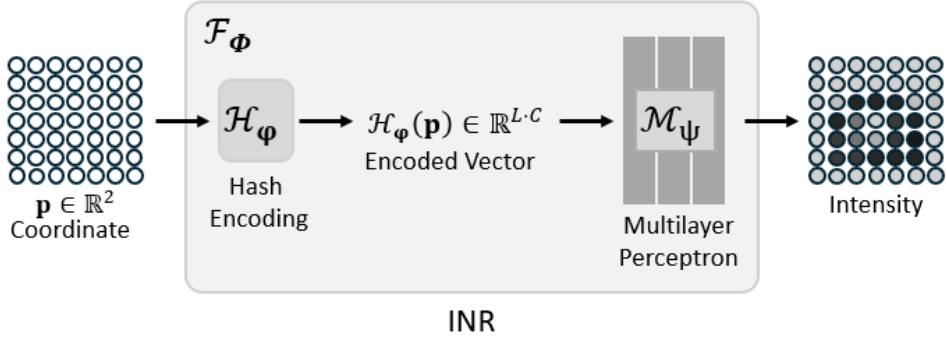

**Figure 8. Detailed architecture of the INR component.** The INR comprises a multi-resolution hash encoding module followed by an MLP. Both components are trainable during the MAR process. The INR maps spatial coordinates to the intensities of the target image, thereby representing the image as a continuous function.

5.1 Problem Formulation

Given the measurement data $\mathbf{y}$ (e.g., metal-affected sinogram), the typical approach is to treat the metal-affected areas as missing data [5, 25]. The MAR problem is then formulated as minimizing an objective function composed of two parts:

where $\hat{\mathbf{x}}$ is the desired artifact-free CT image, $\mathbf{A}$ is the CT forward operator that projects an image to the sinogram domain, and $\mathbf{Tr}$ denotes the binary metal trace indicating the metal-corrupted areas in the sinogram domain.

$$\hat{\mathbf{x}} = \underset{\mathbf{x}}{argmin} \underbrace{\|(1 - \mathbf{Tr}) \odot (\mathbf{A}\mathbf{x} - \mathbf{y})\|^2}_{data\ fidelity} + \underbrace{\mathcal{R}(\mathbf{x})}_{regularization} \quad (1)$$

Here, the data fidelity term encourages the reconstruction result to be consistent with the metal-excluded measurement data $(1 - \mathbf{Tr}) \odot \mathbf{y}$. This term is converted to Eq. 4 and imposed directly by the INR. The regularization R injects the prior knowledge of the CT image distribution to constrain the solution space, which is converted to Eq. 8 and imposed by the DM.

5.2 INR-based Data Fidelity

The data fidelity term in Eq. 1 is imposed by the INR, which is implemented as an MLP preceded by a hash encoding layer, trained to represent a single image. Specifically, the INR maps spatial coordinates to the intensities of the underlying CT image, thereby representing the image as a continuous function. To enable efficient modeling of both low- and high-frequency structures while avoiding the need for extremely deep networks, we adopt a multi-resolution hash encoding layer $\mathcal{H}_\varphi$ that provides a compact yet expressive representation of spatial locations. Therefore, in our implementation, the spatial coordinate $p = (x,y) \in \mathbb{R}^2$ is first transformed via a multi-resolution hash encoding:

$$\mathcal{H}_\varphi: \mathbb{R}^2 \to \mathbb{R}^{L \cdot C}, \quad \mathcal{H}_\varphi(p) = [h_1(p), \dots, h_L(p)] \tag{2}$$

where $L$ is the number of resolution levels and $C$ is the feature dimension per level. The encoded coordinate $\mathcal{H}_\varphi(p)$ is then passed to the MLP $\mathcal{M}_\psi$ to predict the INR-represented image:

$$\mathcal{M}_\psi: \mathcal{H}_\varphi(\mathbf{p}) \in \mathbb{R}^{L \cdot C} \to \mathbf{x}(\mathbf{p}) \in \mathbb{R} \tag{3}$$

where **p** can be any spatial coordinate in a continuous space and **x(p)** is the intensity of the represented image at coordinate **p**. The detailed architecture of the INR is depicted in Figure 8, and we denote the entire INR as $\mathcal{F}_\Phi$, which integrates both the hash encoding layer $\mathcal{H}_\varphi$ and the MLP $\mathcal{M}_\psi$ that follows it.

In this way, we can represent the underlying CT image as a continuous function using an INR network, and subsequently employ line integrals to simulate the CT forward projection process. Compared with directly summing over discretely sampled pixel intensities in the image domain, this formulation more faithfully reflects the underlying physical process of X-ray attenuation. As a result, it allows for a more accurate application of the forward model, thereby ensuring stronger data fidelity in the reconstruction process.

For the MAR problem, the INR-based data fidelity involves updating the network weights $\Phi$ based on the measurement data **y**:

$$\widehat{\Phi}_{fide} = \arg\min_\Phi \frac{1}{|\mathcal{RS}|} \sum_{\mathbf{r} \in \mathcal{RS}} |(1 - \mathbf{Tr}(\mathbf{r})) \odot (\tilde{\mathbf{y}}(\mathbf{r}) - \mathbf{y}(\mathbf{r}))| \tag{4}$$

where RS denotes the set of X-rays, **y(r)** is the observed projection for ray **r**, and **Tr(r)** is the binary metal trace for ray **r**. The predicted projections $\tilde{\mathbf{y}}(\mathbf{r})$ are obtained by applying the differentiable CT forward operator to the MLP-parametrized image:

$$\tilde{\mathbf{y}}(\mathbf{r}) = \sum_{\mathbf{p} \in \mathbf{r}} \mathcal{F}_\Phi(\mathbf{p}) \cdot \Delta \mathbf{p} \tag{5}$$

Benefiting from the inherent continuity prior of INR, we can train the MLP to obtain reconstructed results that reduce some artifacts by optimizing the objective function described in Eq. 4. However, due to the ill-posedness caused by missing sinogram data in metal-affected regions, INR may still produce suboptimal results. To address this, we use a DM to learn the prior distribution of CT images and inject it as a regularization term into the optimization process of the INR. The specific implementation of diffusion-based regularization will be discussed in detail in the next section.

5.3 Diffusion-based Regularization

Inspired by previous works [31, 32], we employ a pre-trained unconditional DM as the regularizer in Eq. 1 and plug it in the INR optimization process. To ensure that the distribution prior learned by the diffusion model regularizes the INR optimization, our approach is to input the content represented by the INR into the diffusion model for enhancement, aligning it with the ideal CT image manifold. We then minimize the distance between the two to update the network parameters of the INR.

Specifically, we first discretize the continuous function implicitly represented by the INR network $\mathcal{F}_\Phi$ into an image $\mathbf{x_0} = \mathcal{F}_\Phi(\mathbf{p})$. Gaussian noise is then injected into $\mathbf{x_0}$ to map it onto the noise manifold at diffusion step $t$ as:

$$q(\mathbf{x}_t \mid \mathbf{x}_0) = \mathcal{N}\left(\mathbf{x}_t; \sqrt{\bar{\alpha}_t}\hat{\mathbf{x}}_0, (1-\bar{\alpha}_t)\mathbf{I}\right) \quad (6)$$

subsequently, by leveraging the Tweedie's formula, we can obtain $\hat{\mathbf{x}}_0$ by performing one step denoising as:

$$\hat{\mathbf{x}}_0 \approx \frac{\left(\mathbf{x}_t - \sqrt{1-\bar{\alpha}_t}\epsilon_\theta(\mathbf{x}_t, t)\right)}{\sqrt{\bar{\alpha}_t}} \quad (7)$$

where $\bar{\alpha}_t = \prod_{i=1}^{t}(1-\beta_i)$, $\{\beta_1, ..., \beta_T\}$ is a fixed variance schedule, and $\epsilon_\theta(\mathbf{x}_t, t)$ is the predicted noise by the DDPM. In summary, by diffusing and denoising the image $\mathbf{x_0}$ represented by the INR, the CT distribution prior learned by the DM is injected into $\hat{\mathbf{x}}_0$.

Finally, we minimize the point-wise loss between the image represented by the INR and the image enhanced by the DM to update the MLP network parameters $\mathbf{\Phi}$, thereby injecting the prior information obtained from the DM into the INR:

$$\hat{\mathbf{\Phi}}_{regu} = \underset{\mathbf{\Phi}}{\operatorname{argmin}} \|\mathcal{F}_\Phi(\mathbf{p}) - \hat{\mathbf{x}}_0(\mathbf{p})\|^2 \quad (8)$$

5.4 Implement details

We use the pre-trained DM from [25] for all diffusion-based methods. For INR, we implement the MLP network with one hash encoding layer [14] and two hidden layers of width 64 through the Tiny-CUDA-NN library [29]. The hidden layers use the ReLU activation function, while the output layer employs a linear activation function. The hash encoding is set with 16 feature levels, 8 features per level, and the size of the hash map is $2^{19}$. The network is trained for 1000 epochs during one iteration. The total diffusion process of the DM is discretized into 1000 timesteps. The diffusion timestep t in the one-step denoising process (Eq. 6) is empirically sampled in descending order from 1000 at intervals of 50, so that multi-level structural regularization can be progressively imposed during the INR training process. Owing to the high noise magnitudes injected during the early phase of the diffusion process, the demand for pixel-level accuracy in the corresponding clean image is relaxed. Therefore, in our practical implementation the INR is empirically trained with far fewer epochs than 1000 when t < 200, which can greatly accelerate training without harming fidelity.

ACKNOWLEDGEMENTS

COMPLIANCE WITH ETHICS GUIDELINES

The authors declare that they have no conflict of interest or financial conflicts to disclose.

All procedures performed in studies involving animals were in accordance with the ethical standards of the institution or practice at which the studies were conducted, and with the 1964 Helsinki declaration and its later amendments or comparable ethical standards.